\def\year{2021}\relax
\documentclass[letterpaper]{article} 
\usepackage{aaai20}  
\usepackage{times}  
\usepackage{helvet} 
\usepackage{courier}  
\usepackage[hyphens]{url}  
\usepackage{graphicx} 
\usepackage{graphicx}  
\usepackage{enumerate} 
\usepackage{multirow}
\usepackage{color}
\usepackage{hyperref}
\usepackage{marvosym}

\usepackage[switch]{lineno}

\title{Multi-Level Attentive Convoluntional Neural Network for Crowd Counting}

\author{Mengxiao Tian\textsuperscript{\rm 1}, Hao Guo\textsuperscript{\rm 2}, Chengjiang Long\textsuperscript{\rm 3} \\
\textsuperscript{\rm 1}School of Computer Science \& Technology, Beijing University of Technology, Beijing, China 100022\\
\textsuperscript{\rm 2}College of Land Science and Technology, China Agricultural University, Beijing, China 100083 \\
\textsuperscript{\rm 3}JD Finance America Corporation, Mountain View, CA, USA 94043 \\
mengxiao\_tian@bit.edu.cn, guohaolys@cau.edu.cn, chengjiang.long@jd.com
} 

\begin{document}

\maketitle

\begin{abstract}
Recently the crowd counting has received more and more attention. Especially the technology of high-density environment has become an important research content, and the relevant methods for the existence of extremely dense crowd are not optimal. In this paper, we propose a multi-level attentive Convolutional Neural Network (MLAttnCNN) for crowd counting. We extract high-level contextual information with multiple different scales applied in pooling, and use multi-level attention modules to enrich the characteristics at different layers to achieve more efficient multi-scale feature fusion, which is able to be used to generate a more accurate density map with dilated convolutions and a $1\times 1$ convolution. The extensive experiments on three available public datasets show that our proposed network achieves outperformance to the state-of-the-art approaches.
\end{abstract}

\section{Introduction}
With a wide range of surveillance applications, crowding counting has been investigated continuously in recent years. It can be applied into safety monitoring, crowd estimation, traffic management, disaster relief, and urban planning, etc. However, due to the presence of drastic scale variances, illumination changes, complex backgrounds, and irregular distribution of human beings, crowd counting is still a very challenging task in reality. 

\begin{figure}[t]
	\centering
	\includegraphics[width=0.99\columnwidth]{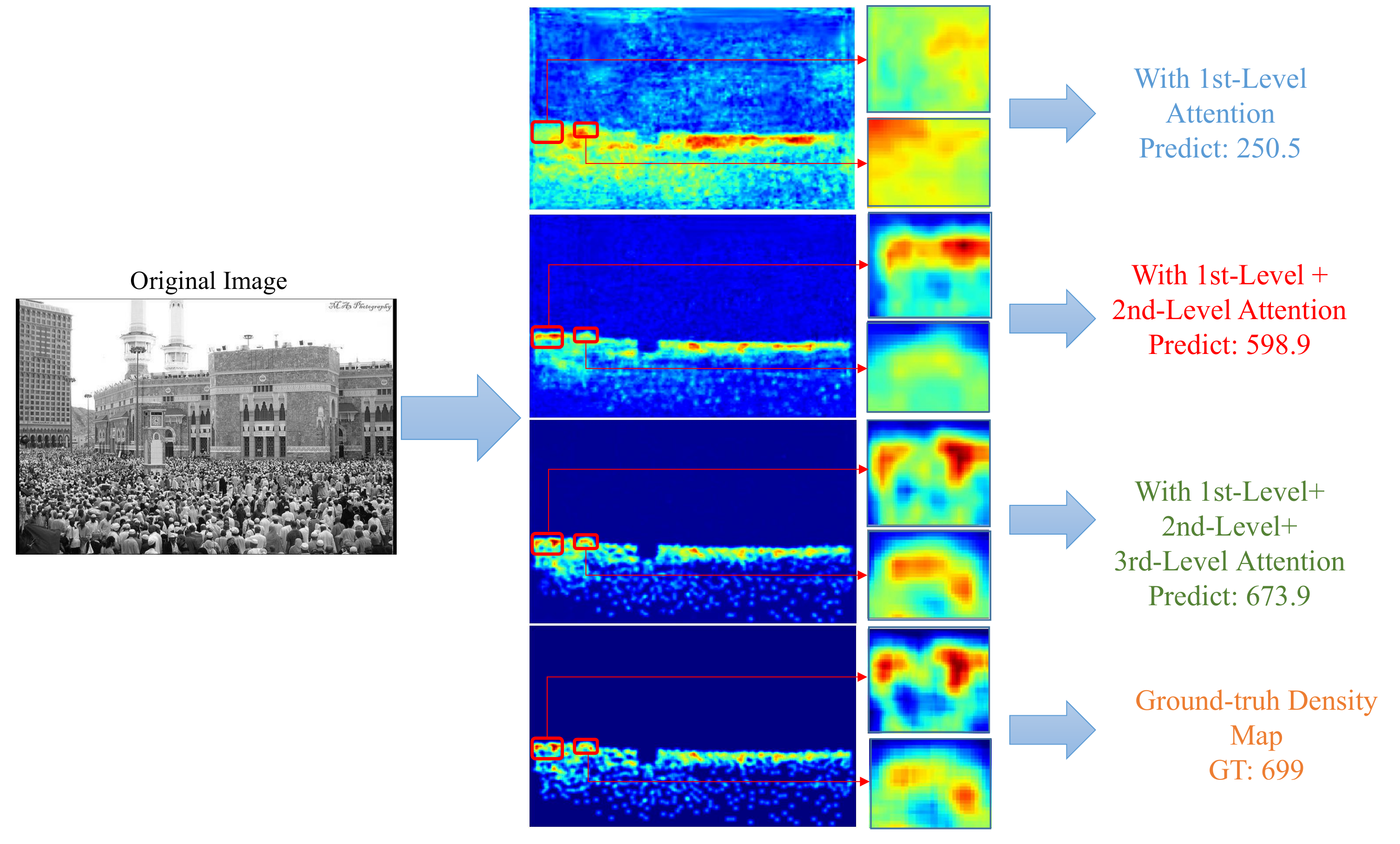} 
	\caption{The visualization of effect on mulit-level attention maps. The prediction number is 250.5, 598.9, and 673.9, from top to bottom. The ground-truth number is 699.}
	\label{intro_sample1}
\end{figure}

Various advanced deep learning techniques based on detection~\cite{Ge2009MarkedPP}, regression~\cite{Idrees2013Multi,ParagiosR01}, and density estimation~\cite{Zhang2016Single,cao2018scale}, have been proposed to deal with this challenge. 
In particular, density estimation based methods have achieved great advances. These techniques have the ability to localize the crowd by predicting a density map as a pixel-wise regression and aggregating to a final counting as the integral of the density map. To generate maps with a retained spatial size as the input, deep Convolutional Neural Network (CNN) architectures~\cite{boominathan2016crowdnet,sam2017switching,Zhang2016Single} are widely applied to handle scale variations by combining different network structures with different receptive field sizes, or fusing multi-context information to formulate a semantic feature representation and facilitate accurate pixel-wise regression. This motives us to further investigate multi-scale and multi-context to exploit the potential of the underlying representation for more accurate pixel-wise estimation.
It is worth noting that attention mechanism~\cite{Ding:ICCV2019}~\cite{Wei:CGF2019}~\cite{Islam:CVPR2020}~\cite{Liu:CVPR2020}~\cite{zhang2020multi}~\cite{Hu:TIP2021}~\cite{Islam:AAAI2021}~\cite{zhang2021two} is able to heighten the sensitivity to features containing the most valuable information. Several attempts have been exerted to incorporate attention mechanism as an effective technique processing into crowd counting~\cite{Jiang2017DecideNet,Zhang2018Attention,kang2018crowd}. However, most of existing research works only use simple attentions and single complicate attention to adjust the model weights or determine the weights to generate a density map at different scales, which can not handle complicated scenes well in the crowd counting problem. This motivates us to further explore multi-level attention mechanism to focus on key pieces of the feature spaces for crowd counting and differentiate irrelevant information. 

In this paper, we propose a multi-level attentive Convolutional Neural Network (MLAttnCNN) for crowd counting, as shown in Figure~\ref{ourmodel}. In order to extract global and sub-regional contextual feature, we introduce the multi-scale spatial pooling with multiple bin sizes, which correspond to different perception fields for capturing different sizes of human heads and overcoming the constraints of fixed sizes of deep networks with full connection layers. Then three-level attention modules are fully explored to extract the contextual feature representation before we produce a density map with dilated convolutions and $1\times1$ convolution. 

We shall emphasize that the 1st-level channel-wise attention module and the 2nd-level spatial attention module are designed to recover more detailed information after upsampling at each scale pooling to formulate more informative feature maps. With multi-scale fusion, we get rich characteristics for the feature representation. 
It enables the entire learning model to have the ability to perceive multi-scale targets and incorporate contextual semantic features, which can better preserve the underlying details to the upper level. 

Inspired by the success of channel-wise attention and spatial attention, we design the 3rd-level triplet attention module by exploring channel, row, and column attention, to rescale the intermediate features by apply the channel, row, and column multiplication for further improving the quality of our feature representation. The intuition behind that we believe one-dimensional attention mechanism should be helpful to impose an attention weight to each element on the corresponding dimension, and the full consideration should also cover row and column information. 
The effects of three-level attention maps are illustrated in Figure~\ref{intro_sample1}. We experimentally observe that with the triplet attention, our MLAttnCNN achieves more accurate estimation in crowd counting.

To sum up, our contributions are three-fold:
\begin{itemize}
	\item We propose a multi-level attentive Convolutional Neural Network to enhance its selectivity in spatial and channel information, improving the effectiveness of multi-scale cascade.
	\item We design an elegant triplet attention module which is able to extract contextual feature by exploring the channel, row, a    nd column attention, to further improve the quality of our feature representation.
	\item We experimentally demonstrate that our proposed method achieves excellent performance on three common benchmark datasets, and achieves comparable and even better performance to state-of-the-art approaches.  
\end{itemize}

\section{Related Work}
The related work can be divided into: {\em density map regression for crowd counting} and {\em attention machenism for crowd counting}. 

{\bf Density map regression for crowd counting} starts from~\cite{lempitsky2010learning} in which density maps are generated with the minimum annotation of a single point blurred by the Gauss kernel for the training of counting networks. 
HydraCNN~\cite{O2016Towards} was proposed to solve the multi-scale issue by extracting multiple overlapping image patches from the multi-scale pyramid of each input image. 
\cite{boominathan2016crowdnet} combines deep and shallow fully convolutional network to capture high-level semantic features and address different scale changes in high-density crowd. \cite{sam2017switching} and \cite{Zhang2016Single} use a multi-column network to handle changes in different head sizes in an image. 
These networks add extra computational costs, and inefficient branches do not adapt well to large-scale changes in crowd density, nor do they adequately address multi-scale changes at all levels, nor do they make good use of spatial and semantic features. This causes  the estimated accuracy in intensive scenarios not to meet the needs in the real-world applications.

{\bf Attention mechanism for crowding counting} is popular in recent works.  
\cite{Jiang2017DecideNet} uses attention mechanism to adjust the model weights adaptively according to the change of crowd density. 
\cite{Zhang2018Attention} proposes the attention mechanism to estimate the probability map and use the predicted probability map for the non-head region of the density estimation. \cite{kang2018crowd} inputs each scale with the final feature map and gets an attention map, multiplies the softmax results of each scale attention map with the density maps of each scale, and finally fuses all density maps with 1$\times$1 convolution. \cite{Hossain2019Crowd} achieves dense crowd counting by a global attentional network with local scale perception. \cite{zhang2019multi} generate a score map using a multi-resolution attention model in which the head position has a higher score that the network structure can pay attention on the head areas in a complex background suppressing the non-head areas.

In this work, we utilize multi-scale pooling module with multi-level attentions to choose relevant and important multi-scale features. We employ the dilated convolutions~\cite{Yu2016Multi,li2018csrnet,2019arXiv190609707D} which have been proven to provide more global multi-scale sensing information without the use of multiple inputs or complex networks. 

\section{Methodology}
\begin{figure*}[t]
	\centering
	\includegraphics[width=0.99\textwidth]{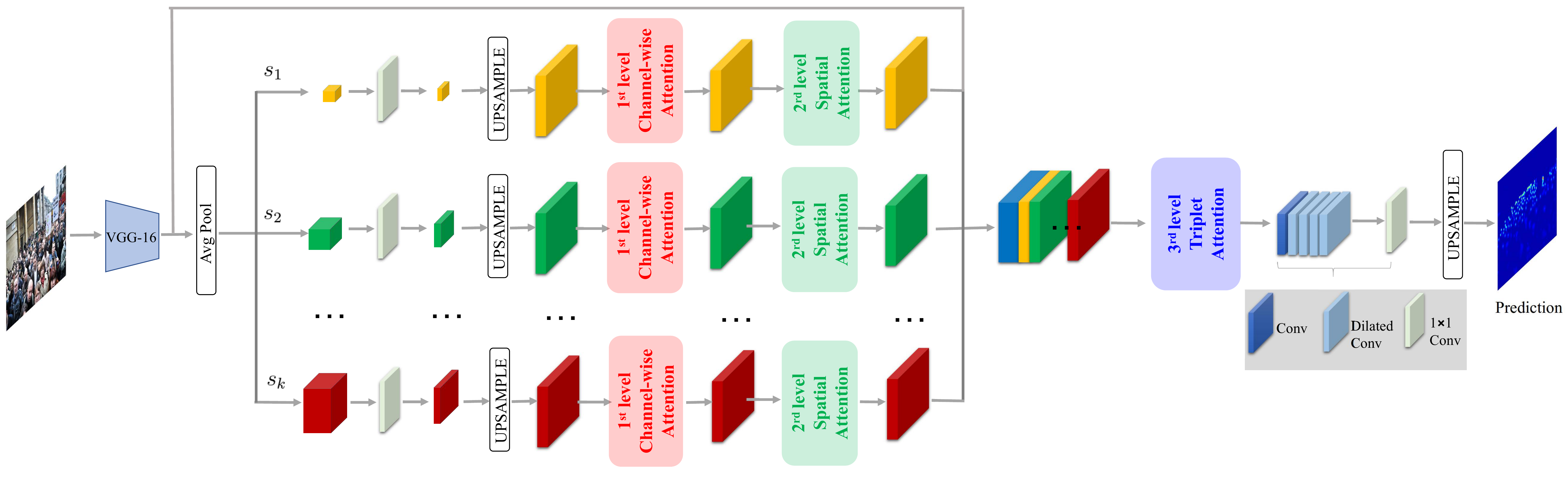} 
	\vspace{-0.3cm}
	\caption{The pipeline of our proposed multi-level attentive convolutional neural network for crowd counting. It consists of an VGG-16 backbone as feature extraction, multi-scale average pooling, 1st level channel-wise attention, 2nd level spatial attention, 3rd triplet attention, and dilated convolutions and $1\times1$ convolutions to generate a density map.}
	\label{ourmodel}
\end{figure*}

As shown in Figure~\ref{ourmodel}, the pipeline of our proposed multi-level attentive convolutional neural network contains four modules, {\em i.e.}, feature extraction with an VGG-16 backbone, multi-scale pooling module, the 1st level channel-wise attention module, the 2nd level spatial attention, the 3rd triplet attention, and dilated convolutions and $1\times 1$ convolution to generate a density map. 

Given an input image, we employ the VGG-16 network~\cite{simonyan2014very} structure as the backbone of extraction feature. Note that we remove three pooling layers to preserve larger feature map, which reduces the loss of spatial information. The final feature map size is 1/4 of the input image, which fed into the multi-scale average pooling module to obtain context information. In this paper, we use five different pooling scales with the sizes of 1$\times$1, 3$\times$3, 5$\times$5, 7$\times$7, 9$\times$9. Each of them has the same number of channels. Then the multi-scale features are upsampled to the same size and fed into the 1st-level channel-wise attention module and the 2nd-level spatial attention for further processing. All the scale features obtained after applying the first two level attention modules are concatenated together with the original feature map to form the contextual feature representation, which is then fed into the 3rd-level triplet attention module, and finally fed into a series of dilated convolution layers and $1\times 1$ convolution layer before the density map is generated.

After obtaining five feature maps in different scales, merging them may  not be the most effective method. The scale of each object varies according to the location of the image because of the scene perspective. Especially for the scale changes of people head, the direct fusion easily loses the spatial information which causes blurring. Therefore, each position of the feature map needs different fusion weights. So we employ multi-level attentions to estimate the attention map for each scale, which could enrich information and more efficient multi-scale integration. 

We are going to discuss three levels of attention module with implementation details as follows.

\subsection{The 1st-Level Channel-wise Attention Module}
\begin{figure}[ht!]
    \vspace{-0.3cm}
	\centering
	\includegraphics[width=0.99\columnwidth]{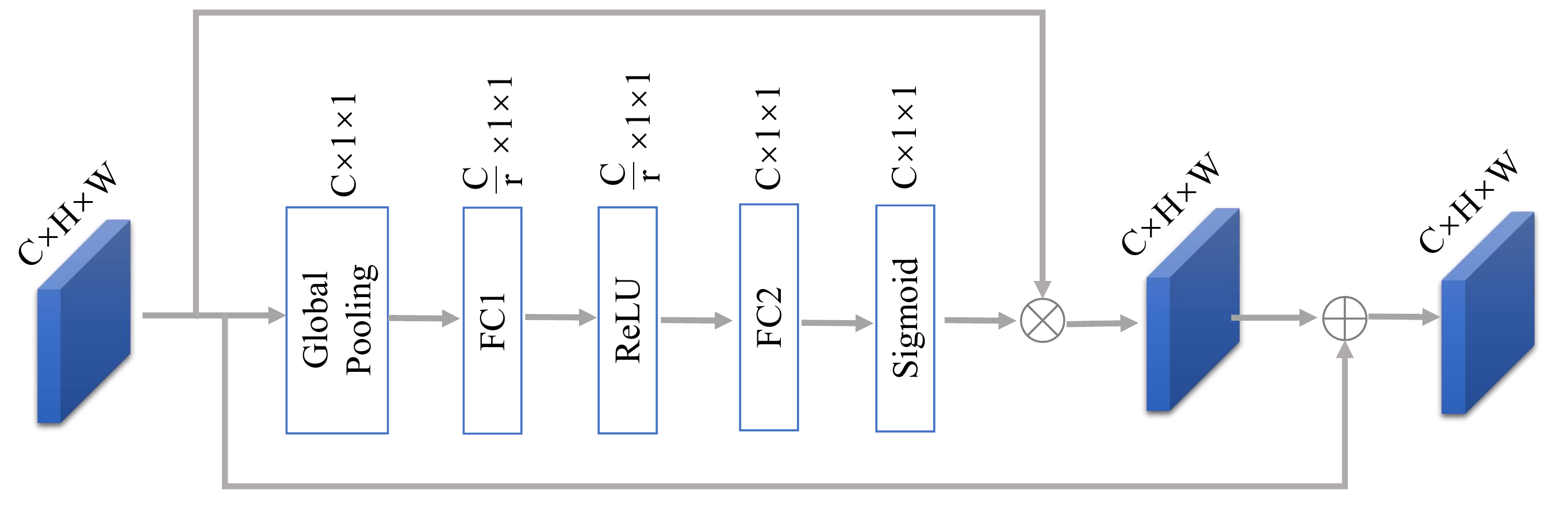}
	\vspace{-0.3cm}
	\caption{The 1st-level channel-wise attention module.}
	\label{fig:1stAttn}
	\vspace{-0.3cm}
\end{figure}

We used a channel-wise attention module based on the high-level feature map as input and generated a channel attention map, and then used to launch the feature map along the channel dimension, as illustrated in Figure~\ref{fig:1stAttn}. For a convolutional feature map $\mathbf{X}\in\mathbf{R}^{H \times W\times C}$, we first squeeze it in spatial dimension by using global avg pooling to get feature map with 1$\times$1$\times$C size, which is then followed by two FC layers and a Sigmoid layer. The first FC layer obtains feature vector with 1$\times$1$\times$ $\frac{C}{r}$ size, where C is the number of channels, and r is scaling parameters. The purpose of r is to reduce the number of channels and reduce the amount of computation. And then followed by the ReLU activation, the dimension of its output remains unchanged. The output dimension of the second FC layer is 1$\times$1$\times$C. Finally, we get the feature vector $\mathbf{S}\in\mathbf{R}^{1 \times 1\times C}$ by using Sigmoid, and final channel-wise attention map by making an element-wise multiplication between $\mathbf{S}=Sigmoid(FC_{2}(ReLU(FC_{1}(\mathbf{X}))))$ and $\mathbf{X}$ for generation. We merge original CNN feature map and channel-wise attention map as the input for the 2nd-level spatial attention module.

\subsection{The 2nd-Level Spatial Attention Module}
\begin{figure}[ht!]
    \vspace{-0.3cm}
	\centering
	\includegraphics[width=0.99\columnwidth]{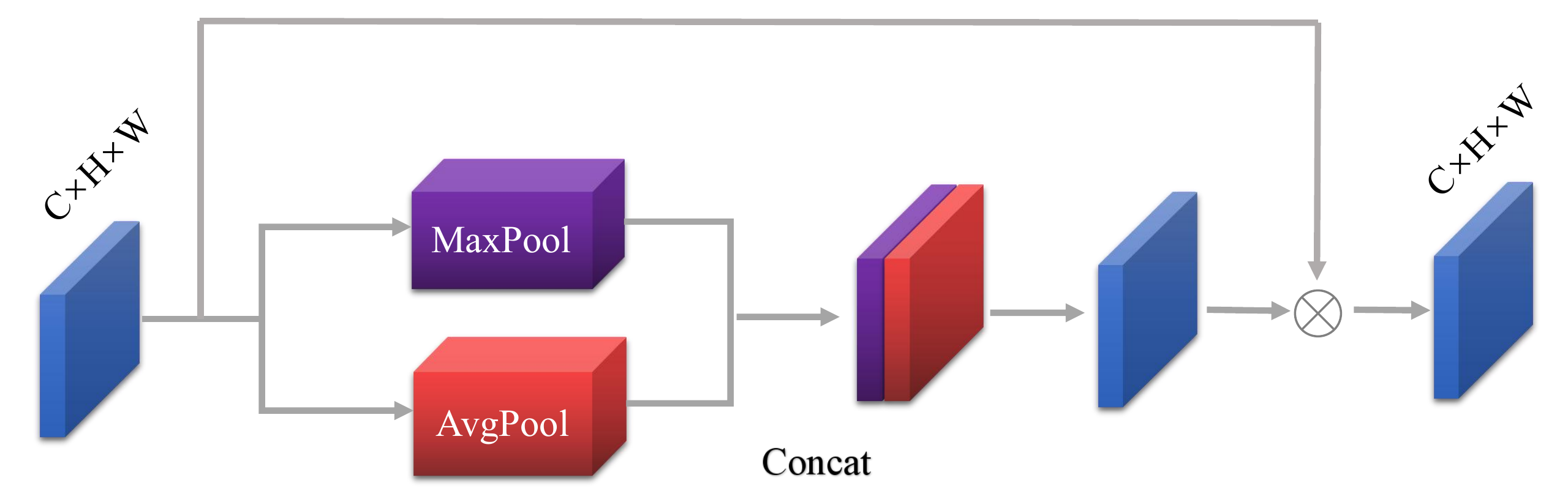}
	\vspace{-0.3cm}
	\caption{The 2nd-level spatial attention module.}
	\label{fig: 2ndAtten}
	\vspace{-0.3cm}
\end{figure}

We put the output channel-wise attention maps focus on spatial attention. Different from the 1st-level channel-wise attention that enhances the correlation between objects and image captions, the 2nd-level spatial attention is designed to focus on location information, which selects attentive areas to enhance the response of feature map, as shown in Figure~\ref{fig: 2ndAtten}. We adopt the spatial attention module \cite{Woo2018CBAM} and make an element-wise multiplication between channel-wise attention map and spatial attention map to actuate the feature map before concating with the original CNN feature. Firstly, it employs the max pooling and avg pooling on channel dimensions to get different feature vectors $\mathbf{X}_{max}\in\mathbf{R}^{1\times H \times W}$ and $\mathbf{X}_{avg}\in\mathbf{R}^{1\times H \times W}$ respectively, and we merge these two feature vector by using concatenation. Finally generate the spatial attention map $\in\mathbf{R}^{H\times W}$ by the convolution operation, and each spatial layer along with batch normalization (bn) layer and ReLU.

\subsection{The 3rd-Level Triplet Attention Module} 
\begin{figure*}[ht!]
	\centering
	\includegraphics[width=0.99\textwidth]{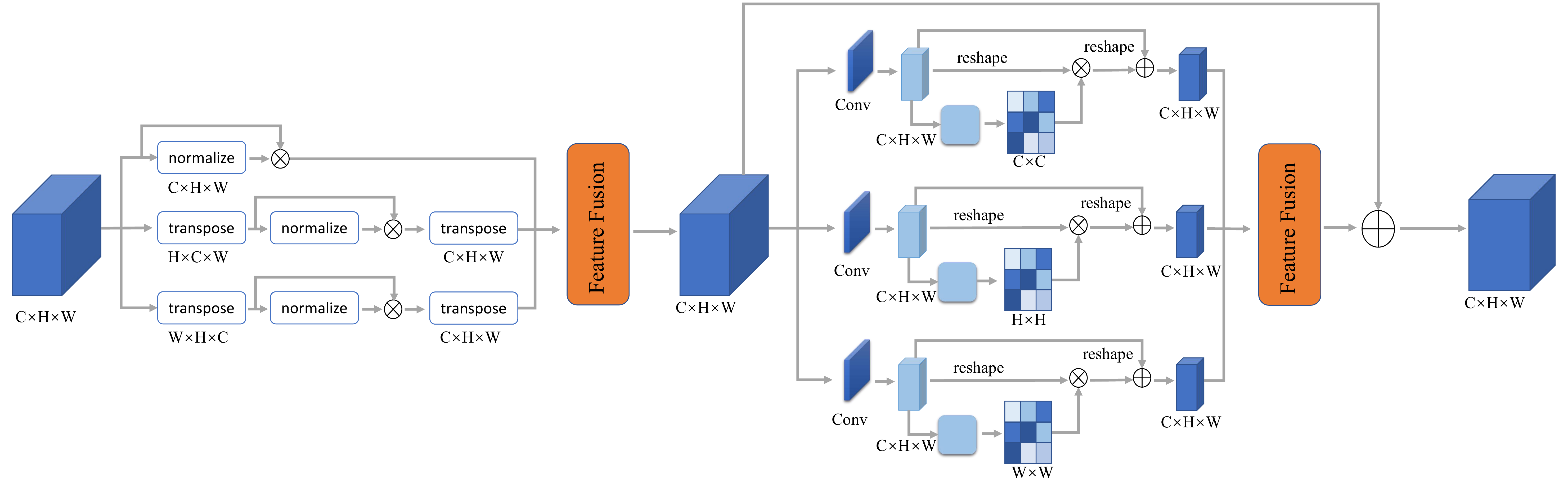}
	\vspace{-0.5cm}
	\caption{The 3rd-level triplet attention module.}
	\label{fig:3rdAttn}
\end{figure*}

We concatenate the five multi-scale actuated attention maps with original CNN features to generate the readjusted density map. 
\color{black}{and} as the input to the 3rd-level triplet attention, 
As demonstrated in Figure~\ref{fig:3rdAttn}, our 3rd-level triple attention has three branches, which is composed of a channel-attention ${\bf \alpha}_1 \in \mathbf{R}^{C\times C}$, a row-attention ${\bf \alpha}_2 \in \mathbf{R}^{H\times H}$, and a column-attention ${\bf \alpha}_3 \in \mathbf{R}^{W\times W}$. We execute the three branch's attention mechanisms respectively to get three dimension feature maps.

The aim of the channel attention is to perform feature recalibration in a global way where the per-channel, per-row and per-column summary statistics are calculated and then used to selectively emphasize informative feature-maps as well as suppress
useless ones ({\em e.g.} redundant feature-maps). 
We first normalize the feature map with a sigmoid activation and multiply it with the original feature map, and then perform two conversions from the input feature map ($X \in \mathbf{R}^{H\times C\times W}$ and $X \in \mathbf{R}^{W\times H\times C}$).
We multiply these two feature maps with the further normalized feature maps. Then we perform transpose to restore the original shape. Finally, we merge the three feature maps together to obtain the normalized feature map.

The normalized feature map first obtains three feature maps through three convolutional layers. We reshape the first feature map $A \in \mathbf{R}^{C\times N}$, where $N=H\times W$, and multiply the transposition of A and reshaped A, and get the channel attention map $A^{'}\in \mathbf{R}^{C\times C}$ by softmax. Each element of the feature map $A^{'}$ is as follows:
\begin{equation}
    A^{'}_{ji} = \frac{exp(A_i\cdot A_j)}{\sum_{i=1}^C{exp(A_i\cdot A_j)}}
\end{equation}
where $A^{'}$ indicates the effect of the feature of the $i^{th}$ channel on the $j^{th}$ channel. We then multiply channel attention map $A^{'}$ and the transpose of A by matrix, multiply the result by a scale factor. Finally we reshape it to the same shape of feature map A and add these two feature maps to get the final output feature map with selective channels $D_c$:
\begin{equation}
    D_{c} = \beta{\sum_{i=1}^C{(A^{'}_{ji}+A_j)}}
\end{equation}

Where $\beta$ is initialized to 0 and gradually learns to assign to a larger weight. Similarly, we can obtain the row-attention ${\bf \alpha}_2$ and the column-attention ${\bf \alpha}_3$ to rescale the input feature with a row multiplication to get the row attention map $\in \mathbf{R}^{H\times H}$, and repeat the previous multiplication and addition to obtain the recalibricated output $D_H$, and with a column mulitplication to get the column attention map $\in \mathbf{R}^{W\times W}$, and repeat the previous multiplication and addition the recalibricated output $D_W$. It is worth mentioning here that we can transpose both the row-attention and the column-attention into the form of channel attention, and then transpose back flexibly. 

The three outputs are summed to obtain the final feature representation:
\begin{equation}
    M_{F} =  a{D_C} + b{D_H} + c{D_W} + X
\end{equation}
where $M_{F}$ is the final feature map we merge three attention maps by three branches, and $a$, $b$ and $c$ represents the weight of each attention map from three branches.
As shown in Figure~\ref{fig:loss_with_hyperparam}, there are the validation loss curves of 6 groups of different a, b, and c training on the UCF\_CC\_50 dataset. When $a=0.8, b=0.15, c=0.05$, the overall model is more convergent. 




Finally, the merged feature map followed by convolution kernel with 3$\times$3 which contains bn layer, we construct four dilated convolution layers with parameters are set to 2, and their all kernel sizes are set to 3$\times$3 to reduce the complexity of network structure. The entire network adopts ReLU as an activation function. In order to map the feature map to the density map, we use a 1$\times$1 filter to produce the final predicted density map, and upsample it to the 1/4 size resolution of the original image.

\subsection{Implementation Details and Loss Function}
In order to supervise our regression model, we use Euclidean distance to measure the distance between the estimated density map and the ground-truth density map. The loss function can be defined as:
\begin{equation}
L(\theta)=  \frac{1}{2N}\sum_{i=1}^{N}\left \| D_i^{gt}-D_i^{pred} \right\|^2,
\end{equation}
where $N$ represents the number of training image patches, $D_i^{gt}$ is the ground-truth map and $D_i^{pred}$ is the estimated density map by our regression module.
The $\theta$ is the parameters of our model learning and $L(\theta)$ learns the loss between the ground-truth density map and the estimated density map. The $D_i^{gt}$ is the density distribution of objects in images can be computed as below:
\begin{equation}
\forall p\in I_i, D^{gt}_i(p)=
\sum_{P\in\mathbf{P}_i}\mathcal{N}(p;P,\sigma^2\mathbf1_{2\times2})
\end{equation}
where $\mathcal{N}(p;P,\sigma^2\mathbf1_{2\times2})$ represents two-dimensional Gaussian distribution, $p$ is the annotated position, and the number of covariance matrix is $\Sigma$. With this density map $D^{gt}_i$, the number of people $N_i$ can be computed as below:
\begin{equation} 
N_i =  {\sum_{p\in\mathbf{P}_i}}D^{gt}_i(p)
\end{equation}
Note that the number of people $N_i$ obtained by integrating the pixels of the predicted density map.

At the training stage, we initialize the end-to-end training model with some parameters of the network using the pre-trained VGG model, and use Adam optimizer~\cite{Kingma2014Adam} in training. For training the whole network, the learning rate is set as 0.0001, and the batch size is 1 in UCF\_CC\_50 dataset, and other three benchmark dataset which use the batch size with 6. 

At the testing stage, we feed the whole input image into the network instead of extracting image patches from the original input image. Finally, we implemented our approach based on Pytorch framework~\cite{paszke2017automatic}.

\section{Experiments}
We evaluate our proposed method on three different crowd counting datasets, {\em i.e.}, UCF\_CC\_50 dataset ~\cite{Idrees2013Multi}, ShanghaiTech dataset ~\cite{Zhang2016Single}, and UCF-QNRF ~\cite{Idrees2018Composition}. 

The {\bf UCF\_CC\_50} dataset is a very challenging dataset. For the distribution of crowd density varies considerably, there are serious problems with occlusion and rapid changes in the number of people per image, which makes it extremely difficult. More specifically, the dataset 
has only 50 images with a total of 63,974 head center annotation provided. The number of head counts vary from 84 to 4543 (1280 in average) per image. Following~\cite{Idrees2013Multi}, we conduct five-fold cross validation and report average test performance.

The {\bf ShanghaiTech Part-B} dataset is a largest open crowd counting dataset in term of the number of annotated people. It contains 716 images with the fixed size of $768\times 1024$ taken from busy streets, covering 88,488 people in total with head center annotations provided. Each image is with the number of people ranging from 9 to 578. Following~\cite{Zhang2016Single}, we take 400 images for training and the rest 316 for evaluation.


The {\bf UCF-QNRF} dataset is a new crowd counting dataset consisting of 1,525 images with a total of 1.25 million annotations, of which crowd counts between 49 and 12,865. It is splitted into training and testing subsets consisting of 1201 and 534 images, respectively.


Note that We fixed the covariance matrix of the Gaussian function to generate the ground-truth density map to ${\Sigma} = 15\cdot{\mathbf1_{2\times2}}$ in UCF\_CC\_50, ShanghaiTech Part-B, and UCF-QNRF dataset. 
For all benchmark datasets, we follow the data augmentation and data processing techniques used in the method~\cite{gao2019c}.

Regarding the evaluation metrics, we adopt Mean Absolute Error (MAE) and rooted Mean Square Error (MSE), which are defined as
\begin{equation}
\footnotesize
\mathrm{MAE} = \frac{1}{\mathit{N}}\sum_{i=1}^{N}{|y_i-\hat{y_i}|} \\
\mathrm{MSE} = \sqrt{\frac{1}{{N}}\sum_{i=1}^{N}(y_i-\hat{y_i})^2}
\end{equation}
where $\mathit{N}$ represents the number of images in test set, $y_i$ represents the actual count in the $\mathit{i}$th image, and $\hat{y_i}$ represents the predicted count in the $\mathit{i}$th image.
MAE measures the accuracy of crowd counting algorithm, MSE measures the robustness of crowd counting algorithm.




\subsection{Comparison with State-of-the-art}
We compared our proposed MLAttnCNN with a series of recent advanced approaches, {\em i.e.},
MCNN~\cite{Zhang2016Single}, 
Cascade-CNN~\cite{sindagi2017cnn}, 
Switch-CNN~\cite{sam2017switching}, 
AT-CSRNet~\cite{zhao2019leveraging}, 
CL-CNN~\cite{Idrees2018Composition}, 
DRSAN~\cite{Liu2018Crowd}, 
CSRNet~\cite{li2018csrnet},
TEDNet~\cite{Jiang2019Crowd}.
SANet~\cite{cao2018scale}, 
TDF-CNN~\cite{sam2018top}, 
ASD~\cite{wu2019adaptive}, 
SL2R~\cite{liu2019exploiting}, 
PACNN~\cite{shi2019revisiting}, 
and CAN~\cite{Liu_2019_CVPR}. 

For fair comparison, we train each model on the same training set with the same data augmentation tricks, 
and evaluate on the same testing set.

\subsubsection{UCF\_CC\_50 dataset}
We summarize the quantitative results in Table~\ref{table2}, from which we can clearly observe that our approach outperforms the state-of-the-art methods and has achieved the best results with MAE of 200.8. This suggests that our proposed method is able to achieve good performance even though the training data is insufficient. 
\begin{table}[ht!]
	\centering
	\begin{tabular}{l|c|cc}
	    \hline
		Method & Venue/Year & MAE & MSE \cr
	    \hline
		DRSAN & IJCAI/2018 & 219.2 & 250.2\\
		SANet & ECCV/2018 & 258.4 & 334.9\\
		TDF-CNN & AAAI/2018 & 354.7 & 491.4\\
		SL2R & PAMI/2019 & 279.6 & 408.1\\
		PACNN & CVPR/2019 & 241.7 & 320.7\\
		CAN & CVPR/2019 & 212.2 & \textbf{243.7} \\
		\hline
		MLAttnCNN & AAAI/2020 & \textbf{200.8} & 273.8\\
	    \hline
	\end{tabular}
    \caption{Estimation errors on the UCF\_CC\_50 dataset.}
	\label{table2}
\end{table}
\begin{figure}[ht!]
    \vspace{-0.5cm}
	\centering
	\includegraphics[width=1.0\columnwidth]{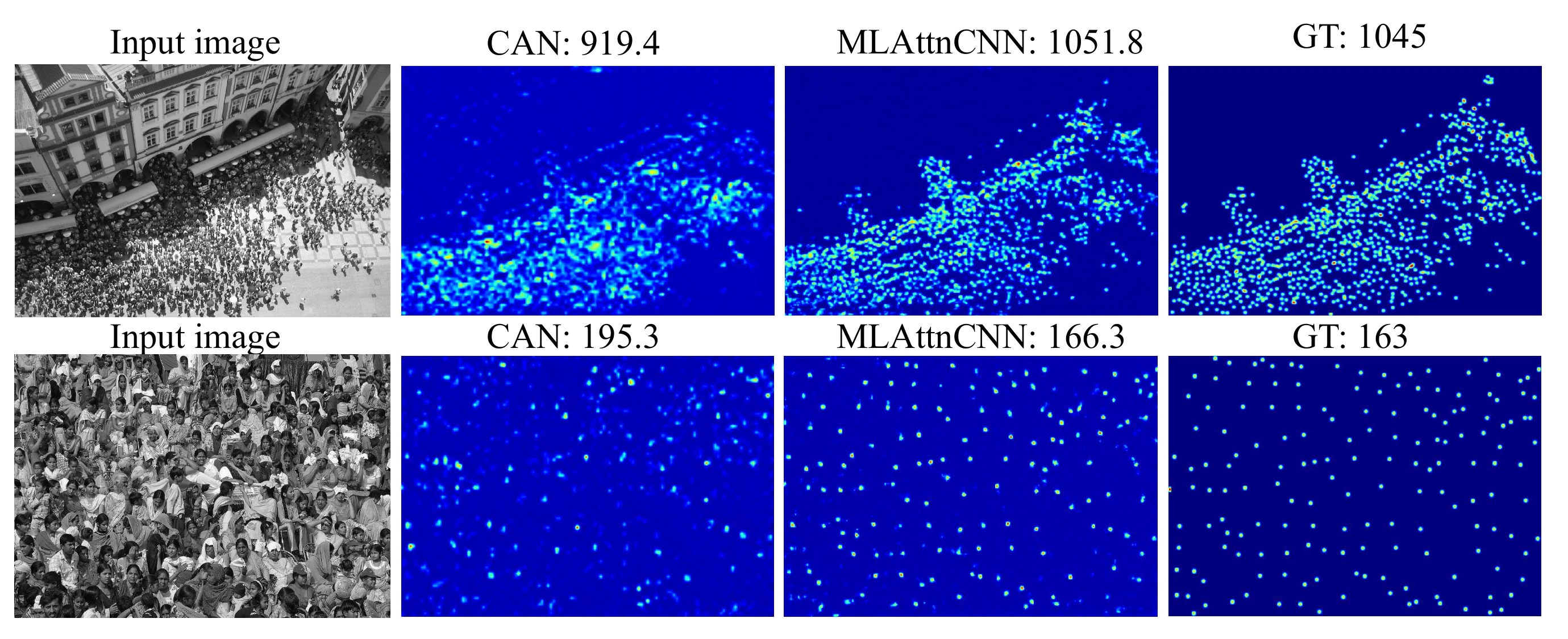}
	\vspace{-0.5cm}
	\caption{Visualization of the predicted density maps on the UCF\_CC\_50 dataset.}
	\label{fig:vis_ucfcc50}
	\vspace{-0.3cm}
\end{figure}


\begin{figure}[ht!]
	\centering
	\includegraphics[width=0.45\textwidth]{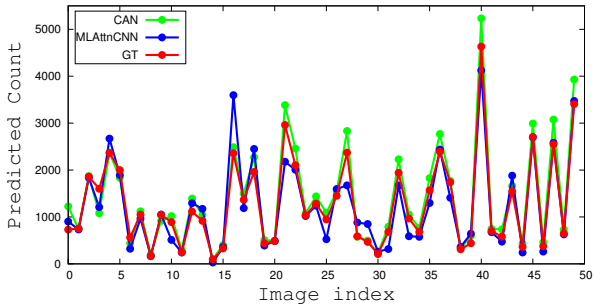} \\ 
	\vspace{-0.5cm}
	\caption{Comparisons of estimated count in UCF\_CC\_50 dataset.}
	\label{fig:statistic_ucfcc50}
	\vspace{-0.3cm}
\end{figure}


For better understand the advantage of our proposed approach, we visualize some examples of density maps estimated on the UCF\_CC\_50 dataset in Figure~\ref{fig:vis_ucfcc50}. We can see that our predicted density map has high quality and corresponding crowd images, and the predicted density maps closely follow the actual crowd distribution in the visualization.

We also compared the predicted and actual counts in the UCF\_CC\_50 dataset, as shown in Figure~\ref{fig:statistic_ucfcc50}. Our method is superior to CAN across most data splits, which improves the robustness and effectiveness of our proposed method.
 For most images, the predicted count is close to the actual count relatively, but while the crowd density is particularly large of an image, the prediction error is also relatively large. We believe that the reason for this may be that the training data is not enough, and lack of crowded image data, resulting in our model not able to learn the image features in training well.
 

\subsubsection{ShanghaiTech Part-B dataset}
\begin{figure}[ht!]
	\centering
	\includegraphics[width=0.45\textwidth]{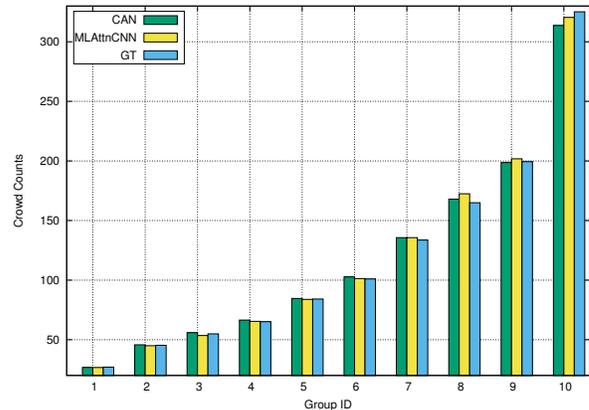} \\ 
	\vspace{-0.5cm}
	\caption{Comparisons of average counting estimates of 10 splits in ShanghaiTech Part-B data set based on the increase of the number of people in each image.}
	\label{fig:vis_shanghaitech_table}
	\vspace{-0.3cm}
\end{figure}

\begin{figure}[ht!]
	\centering  
	\includegraphics[width=0.5\textwidth]{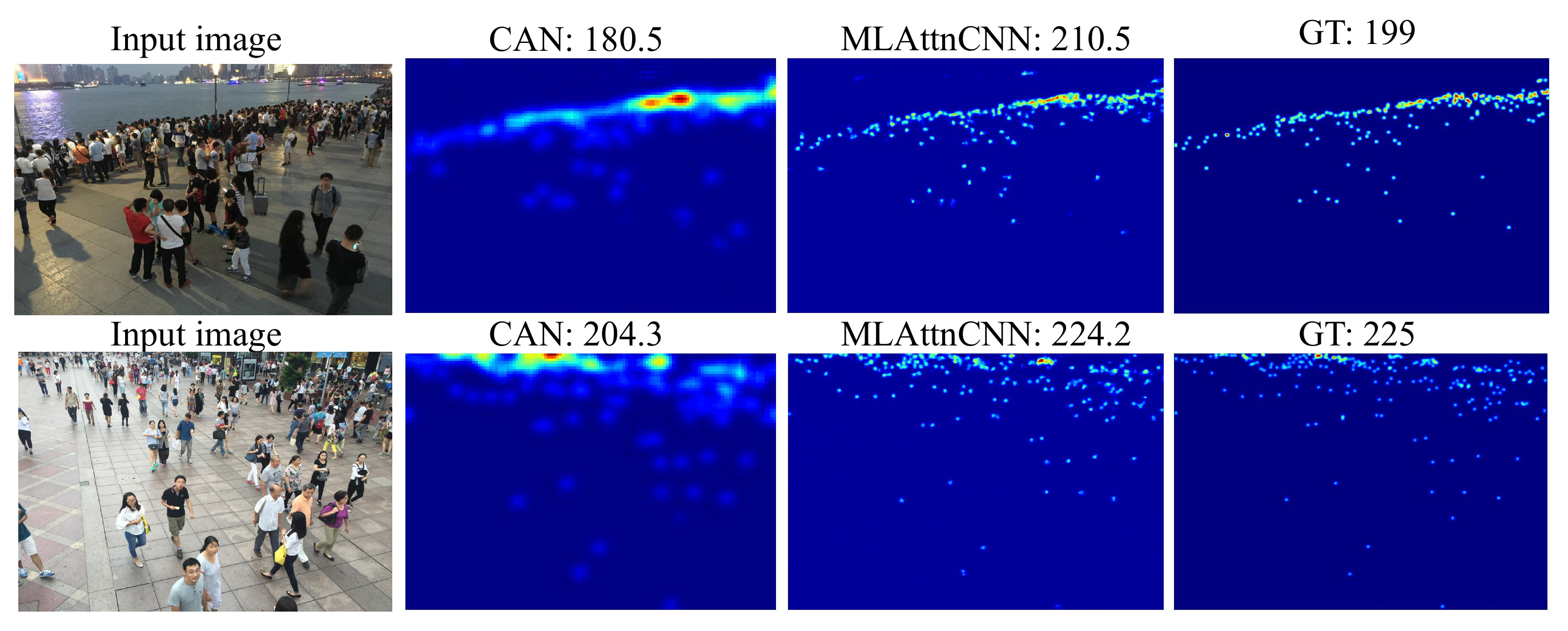} \\ 
	\vspace{-0.25cm}
	\caption{Comparison in visualization on density maps in the ShanghaiTech Part-B dataset.}
	\label{fig:vis_shanghaitech}
	\vspace{-0.3cm}
\end{figure}

Table~\ref{tab:shanghaitech} presents all the quantitative results. As we can observe, compared with other methods, 
we achieve the best MAE of 7.5 and MSE of 11.6. We also provide the visualization results in Figure~\ref{fig:vis_shanghaitech}, and we divide all the testing images into 10 groups according to basis of the crowd count, and count each group on average. Estimated count comes from the prediction of our network structure. We can see that our method is superior to CAN in data splits, which proves its robustness and effectiveness, which as been seen in Figure~\ref{fig:vis_shanghaitech_table}.

\begin{table}[ht!]
	\centering
	\hspace{-0.5cm}\begin{tabular}{l|c|cc}
	\hline
		Method & Venue/Year & MAE & MSE \cr
	\hline
		MCNN & CVPR/2016 &26.4 &41.3  \cr
		Cascade-CNN & AVSS/2017 &20.0 &31.1 \cr
		Switch-CNN & CVPR/2017 &21.6 &33.4  \cr
		CSRNet & CVPR/2018 &10.6 &16.0 \cr
		SANet & ECCV/2018 &8.4 &13.6  \cr
		AT-CSRNet &CVPR/2019 &8.11 &13.53 \cr
		TEDNet & CVPR/2019 &8.2 &12.8  \cr
		CAN & CVPR/2019 & 7.8 & 12.2   \\
	\hline
		MLAttnCNN & AAAI/2020 &\textbf{7.5} &\textbf{11.6}  \cr
	\hline
	\end{tabular}
    \caption{Estimation errors on the ShanghaiTech Part-B dataset.}
	\label{tab:shanghaitech}
\end{table}

\subsubsection{UCF-QNRF dataset}
\begin{figure}[ht!]
	\centering
	\includegraphics[width=0.5\textwidth]{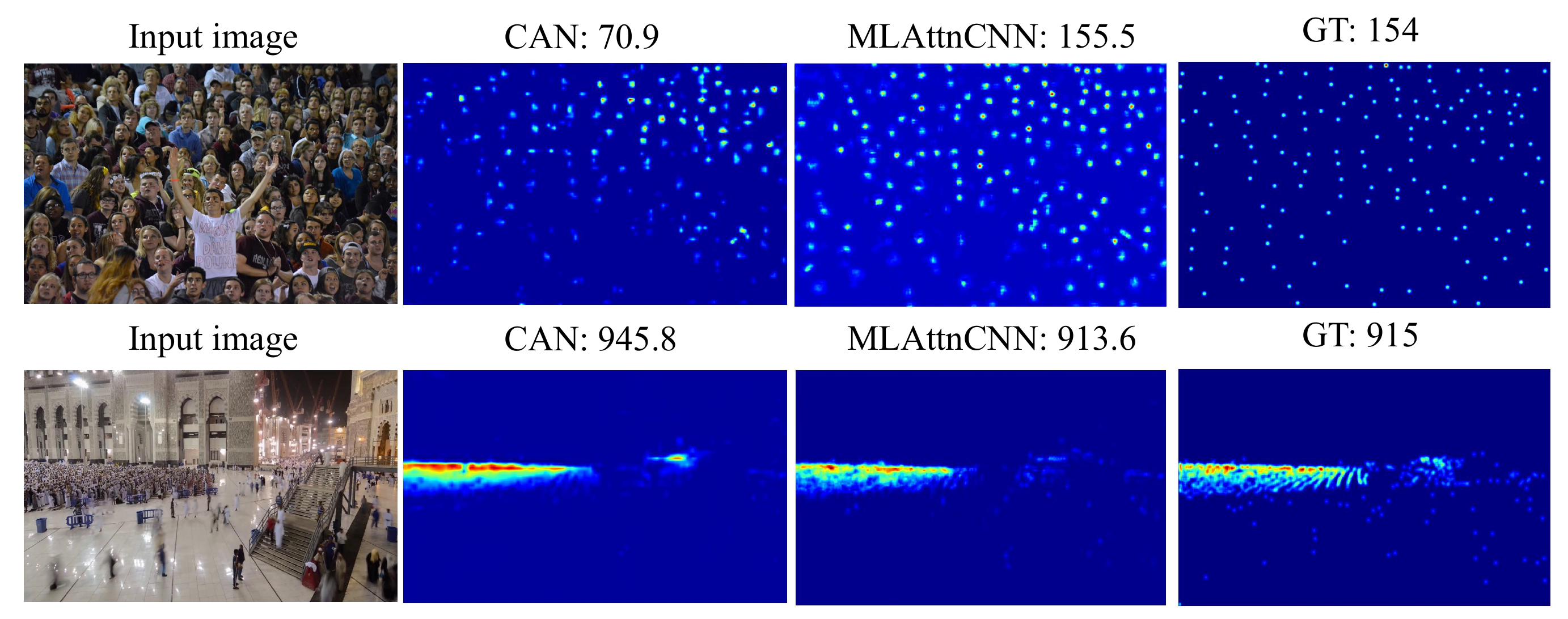} \\
	\vspace{-0.25cm}
	\caption{Visualization of the predicted density maps on the UCF-QNRF dataset.}
	\label{fig:vis_ucfqnrf}
	\vspace{-0.3cm}
\end{figure}

We compare our proposed MLAttnCNN with eleven baselines and report the results in Table~\ref{tab:ucf-qnrf}. On this dataset, we achieve the best MAE of 101.2 and MSE of 174.6. We also visuailize some examples on these two datasets on Figure~\ref{fig:vis_ucfqnrf}. Again, we can observe that our proposed MLAttnCNN model always improves the counting performance.
\begin{table}[ht!]
	\centering
	\hspace{-0.15cm}\begin{tabular}{l|c|cc}
	
	\hline
		Method & Venue/Year & MAE & MSE\cr
	\hline
		MCNN & CVPR/2016 &277 &426\cr
		Cascade-CNN & AVSS/2017 &252 &514\cr
		Switch-CNN & CVPR/2017 &228 &445\cr
		CL-CNN & ECCV/2018 &132 &191\cr
		TEDNet & CVPR/2019 &113 &188\cr
		CAN & CVPR/2019 & 107 & 183 \cr
	\hline
		MLAttnCNN & AAAI/2020 & \textbf{101} &\textbf{175}\cr
	\hline
	\end{tabular}
    \caption{Estimation errors on the UCF-QNRF dataset.}
	\label{tab:ucf-qnrf}
	\vspace{-0.3cm}
\end{table}


\subsection{Ablation Study} 
We conduct ablation study on the UCF\_CC\_50 dataset to justify the design of our proposed MLAttnCNN.

\subsubsection{Effectiveness of backbone structure}	
We choose the VGG-16 structure as our backbone for initial feature extraction. Previous density counting research~\cite{li2018csrnet,Liu2018Crowd,8658887,sindagi2019ha} generally retrain the first 5 sets of convolutional blocks in VGG, and delete the last two pooling layers with the downsampling coefficient as 8. Since too many pooling layers would lead to the loss of spatial information, we remove three pooling layers and change the downsampling coefficient to 4, in order to avoid this situation and improve accuracy. 
\begin{table}[t]
	\begin{tabular}{l|l|cc}
		\hline
			 & VGG-16 structure &MAE &MSE \\
		\hline
			\multirow{2}{*}{3$\times$Pool} &2$(64,3)$+MP, 2$(128,3)$+MP, &\multirow{2}{*}{233.4} &\multirow{2}{*}{315.6}\\ 
			& 3$(256,3)$+MP, 3$(512,3)$  & & \\
		\hline
			\multirow{2}{*}{2$\times$Pool}  &2$(64,3)$+MP, 2$(128,3)$+MP, &\multirow{2}{*}{200.8} &\multirow{2}{*}{273.8}\\
			& 3$(256,3)$, 3$(512,3)$) & & \\
		\hline
	\end{tabular}
    \caption{Results of the different number of pooling layers in UCF\_CC\_50 dataset.  $k(m,n)$ denotes $k$ convolution layers, each of which has $m$ filters with the kernel size of $n\times n$. ``+MP" represents a max pooling layer.}
	\label{table4}
\end{table}

The results of comparison experiments are reported in Table ~\ref{table4}, which shows that we are able to reduce both MAE and MSE values for our proposed MLAttnCNN by eliminating the number of pool layers, resulting in a more accurate density map and improve the crowd counting performance.

\subsubsection{Effectiveness of Multi-Scale Pooling Module}
\begin{table}[t]
	\centering  
	\hspace{-0.75cm}\begin{tabular}{l|l|cc}
		\hline
			&\multirow{2}{*}{Scales}&\multicolumn{2}{c}{UCF\_CC\_50}\\ 
		     && MAE & MSE \\
			\hline
			{4$\times$avg pooling layers} &MS1 &221.2 &276.6 \\
			{4$\times$avg pooling layers} &MS2 &241.3 &324.0 \\
			{5$\times$avg pooling layers} &MS3 &\textbf{200.8} &\textbf{273.8} \\
			{4$\times$avg pooling layers} &MS4 &244.1 &338.5 \\ 
			\hline
	\end{tabular}
    \caption{Comparison of results on both the UCF\_CC\_5 dataset for varying-size pooling bin sizes.}
	\label{table5}
\end{table}
We studied four varying multi-scale (MS) pooling bin sizes to assemble the model, {\em i.e.},
\begin{itemize}
    \item MS1: 1$\times$1, 2$\times$2, 3$\times$3, and 6$\times$6 avg pooling layers.
    \item MS2: 3$\times$3, 5$\times$5, 7$\times$7 and 9$\times$9 avg pooling layers.
    \item MS3: 1$\times$1, 3$\times$3, 5$\times$5, 7$\times$7 and 9$\times$9 avg pooling layers.
    \item MS4: 1$\times$1, 3$\times$3, 5$\times$5, and 7$\times$7 avg pooling layers.
\end{itemize}
We compared these four options on both the ShanghaiTech dataset and the UCF\_CC\_50 dataset. As shown in table ~\ref{table5}, MS4 performs the best in the UCF\_CC\_50 dataset, and MS3 works the best in the ShanghaiTech dataset.
Also, we observe that odd pooling bin size is more suitable for learning density map of CNN Model than even size pooling kernel. With more and more effective semantic information, odd size pooling kernel could accurately locate the spatial distribution of crowd density and improve the accuracy of crowd counting; For the UCF\_CC\_50 dataset, which has fewer images than SHB and UCF-QNRF, and crowds are more crowded and very variable distribution.  Filters with large receptive fields are more suitable for ground-truth density map with larger heads.	

\subsubsection{Effectiveness of attention modules}
\begin{figure}[ht!]
	\centering
	\includegraphics[width=0.99\columnwidth]{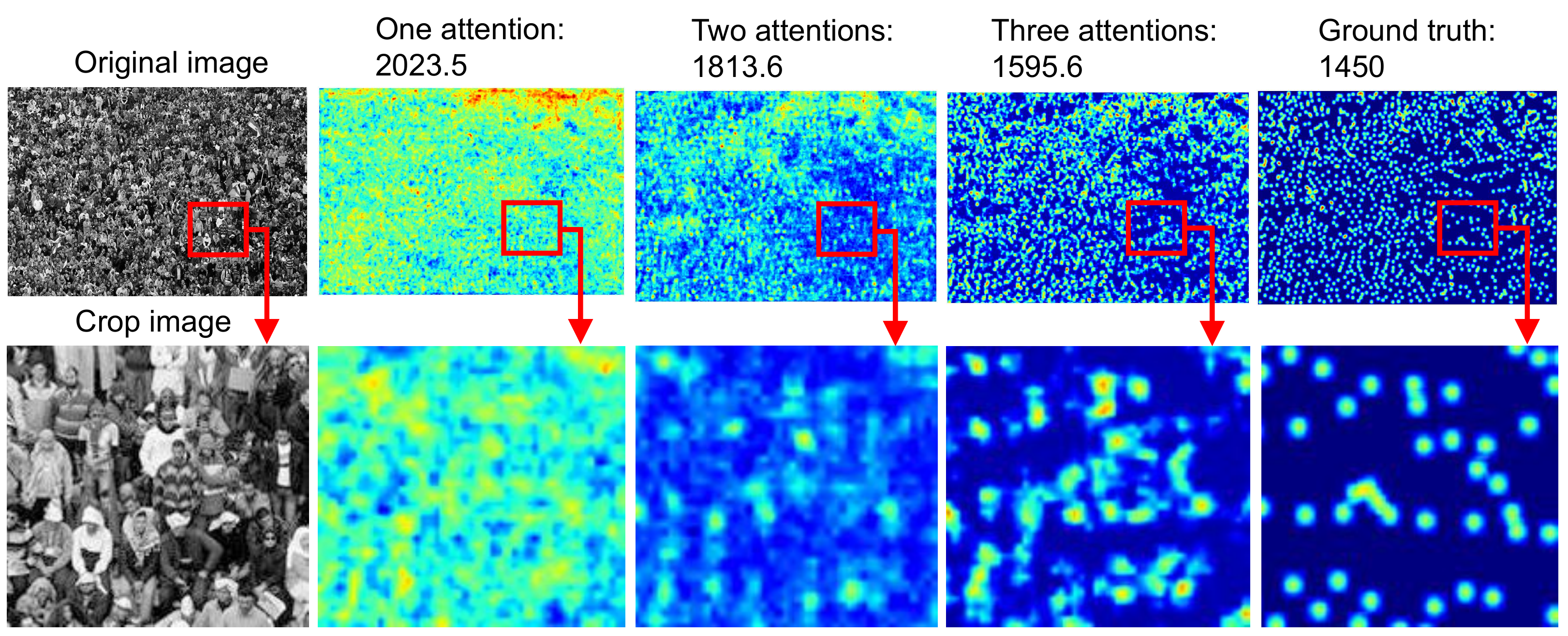} 
	\vspace{-0.25cm}
	\caption{The visualization of effect on mulit-level attention maps. 
	}
	\label{fig:visualize_MLAttn}
\end{figure}

We also study multi-level attentive modules on the UCF\_CC\_50 dataset with three combinations, as shown in  Table~\ref{tab:MLAttn} and Figure~\ref{fig:visualize_MLAttn}. We can see that combining the 1st-level channel-wise attention with the 2nd-level spatial attention works better than use the 1st-level attention only, and the combination of all three-level attentions achieves the best performance.
\begin{table}[ht!]
	\centering
	\begin{tabular}{c|cc}
		\hline
		Methods &MAE &MSE \\
		\hline
		1st-Level Attn &260.5 &333.1\\ 
		1st-Level + 2nd-Level Attn &246.2 &330.3\\ 
		1st-Level + 2nd-Level + 3rd-Level Attn &\textbf{200.8} &\textbf{273.8}\\
		\hline
	\end{tabular}
	\caption{Effect of multi-level attentive modules for crowd counting in our proposed MLAttnCNN on the UCF\_CC\_50 dataset. 
	}
	\label{tab:MLAttn}
\end{table}

\subsubsection{Effectiveness of hyperparameters in 3rd-triple attention}
\begin{figure}[ht!]
  \centering
  \includegraphics[width=0.98\columnwidth]{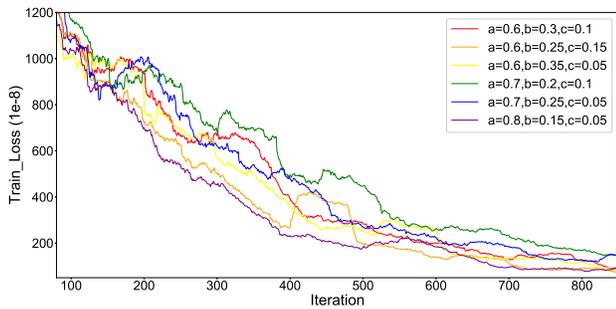}
  \vspace{-0.5cm}
  \caption{The convergence performance for different network  hyperparameters in the UCF\_CC\_50 dataset.}
  \label{fig:loss_with_hyperparam}
  \vspace{-0.2cm}
\end{figure}

We make six different sets of hyperparameter experiments in the UCF\_CC\_50 dataset. 
From the results in 
Figure~\ref{fig:loss_with_hyperparam}, we can observe that our proposed MLAttnCNN achieves the faster convergence and better convergence when we use the hyperparameters $a=0.8, b=0.15, c=0.05$ in the 3rd-level triplet attention.

\section{Conclusion}
In this paper, we propose a multi-level attentive Convolutional Neural Network for crowd counting. We employ multi-scale pooling and multi-level attentions to explore the underlying contextual details for generating a high quality density map. Our approach not only facilitates the semantic perception of the heads of different scales, but also enriches the features of different layers for fusion. It is experimentally proved more suitable for extremely dense people. Extensive experiments strongly demonstrate that our approach is robust and has outperformed the state-of-the-art approaches on multiple crowd counting benchmark datasets.

Our future work includes extending the current work to solve crowd counting problems in videos.

\section{Acknowledgments}
 This research is supported by the National Natural Science Foundation of China [grant numbers 42071449, 41601491]. 

\bibliography{MLAttnCNN}
\bibliographystyle{aaai}
\end{document}